\documentclass[10pt,twocolumn,letterpaper]{article}

\usepackage{cvpr}              %

\usepackage[dvipsnames]{xcolor}

\definecolor{cvprblue}{rgb}{0.21,0.49,0.74}
\usepackage[pagebackref,breaklinks,colorlinks,citecolor=cvprblue]{hyperref}

\def\paperID{} %
\def\confName{3DV\xspace}
\def\confYear{2026\xspace}

\author{
Zicong Fan$^{1,2,3,\dagger}$ 
\quad Edoardo Remelli$^{1}$ 
\quad David Dimond$^{1}$ 
\quad Fadime Sener$^{1}$ \\
\quad Liuhao Ge$^{1}$ 
\quad Bugra Tekin$^{1}$ 
\quad Cem Keskin$^{1}$
\quad Shreyas Hampali$^{1}$
 \\
 {
 $^1$Meta Reality Labs \quad
 $^2$ETH Z{\"u}rich \quad
 $^3$Max Planck Institute for Intelligent Systems, T{\"u}bingen
 }
}

\usepackage{colortbl}

\usepackage{graphicx}
\usepackage{xcolor}
\usepackage{multirow}
\usepackage{float}
\usepackage{tabularx}

\usepackage{microtype}
\usepackage{lipsum}  
\usepackage{bbm}
\usepackage{caption}

\usepackage{times}
\usepackage{epsfig}
\usepackage{amsmath}
\usepackage{amssymb}

\newcommand\blfootnote[1]{%
  \begingroup
  \renewcommand\thefootnote{}\footnote{#1}%
  \addtocounter{footnote}{-1}%
  \endgroup
}

\newcommand{\colorRef}[1]{\textcolor{black}{#1}} %

\newcommand{\refFig}[1]{\mbox{\colorRef{Figure~\ref{#1}}}}

\newcommand{\refTab}[1]{\mbox{\colorRef{Table~\ref{#1}}}}

\newcommand{\refEq}[1]{\mbox{\colorRef{Equation~\ref{#1}}}}

\newcommand{\ccite}[1]{~\cite{#1}}
\newcommand{\subtitle}[1]{\textbf{#1}.}

\definecolor{GreenColor}{rgb}{0.137,0.573,0.565}
\definecolor{OrangeColor}{rgb}{0.914,0.541,0.0.141}
\definecolor{PurpleColor}{rgb}{0.5,0,0.7}
\definecolor{BlueColor}{rgb}{0,0.725,0.949}
\definecolor{PinkColor}{rgb}{0.9843,0.19215,0.6}

\newcommand{\V}[1]{\mathbf{#1}} %
\newcommand{\R}{\rm I\!R}

\newcommand{\norm}[1]{\left\lVert#1\right\rVert}

\DeclareMathOperator*{\argmin}{arg\,min}

\newcommand{\myparagraph}[1]{\noindent\textbf{#1:}}

\newcommand{\nameCOLOR}[1]{\textcolor{black}{#1}} %
\newcommand{\methodname}{\mbox{\nameCOLOR{PALM-Net}}\xspace}

\newcommand{\TITLE}{\datasetname: A Dataset and Baseline for Learning Multi-Subject Hand Prior}

\newcommand{\datasetname}{\mbox{\nameCOLOR{PALM}}\xspace}

\newcommand{\highlightNUMB}[1]{\textcolor{black}{#1}} %

\newcommand{\numSubs}{{\highlightNUMB{$263$}}\xspace}
\newcommand{\numCams}{{\highlightNUMB{$7$}}\xspace}

\newcommand{\numImages}{{\highlightNUMB{$90k$}}\xspace}
\newcommand{\numScans}{{\highlightNUMB{$13k$}}\xspace}

\newcommand{\suppl}{\textcolor{black}{SupMat}\xspace}

\newcommand{\rgb}{RGB\xspace}

\newcommand{\twoD}{{2D}\xspace}

\newcommand{\threeD}{\xspace{3D}\xspace}

\newcommand{\groundtruth}{{ground-truth}\xspace}

\newcommand{\mano}{\mbox{MANO}\xspace}

\newcommand{\interhand}{\mbox{InterHand2.6M}\xspace}

\title{\TITLE} %
\begin{document}
\twocolumn[{%
\renewcommand\twocolumn[1][]{#1}%
\maketitle

\begin{center}
    \centerline{\includegraphics[width=0.90\linewidth]{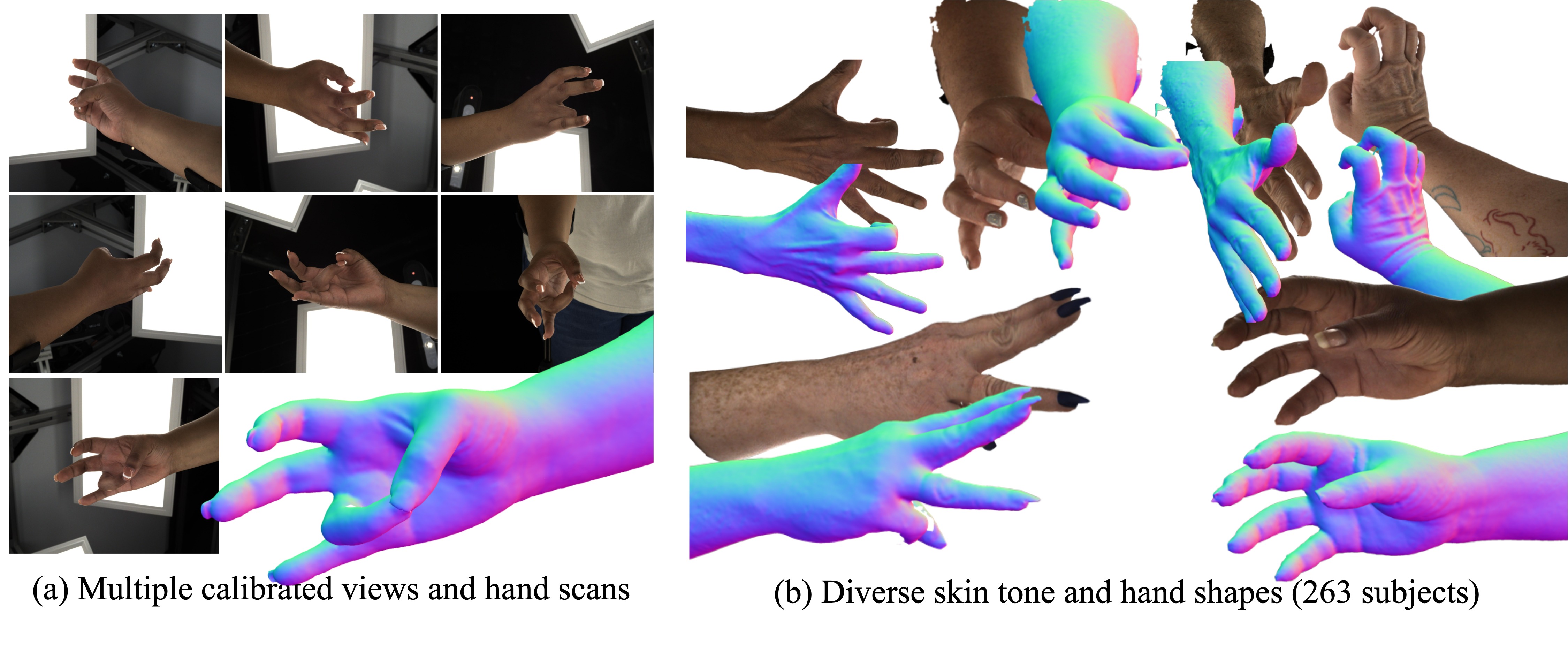}}
    \vspace{-4mm}
    \captionof{figure}{\textbf{Dataset overview:} 
    \datasetname is a large-scale dataset comprising calibrated multi-view high-resolution \rgb images and 3dMD hand scans (a). It features \numSubs subjects spanning a wide range of skin tones and hand sizes, \numImages \rgb images, and \numScans high-quality hand scans with corresponding MANO registrations (b). This diversity and precision provide a foundation for learning a universal prior over human hand shape and appearance.
    }
    \label{fig:teaser}
\end{center}%

}
]
\maketitle

\begin{abstract}

The ability to grasp objects, signal with gestures, and share emotion through touch all stem from the unique capabilities of human hands.
Yet creating high-quality personalized hand avatars from images remains challenging due to complex geometry, appearance, and articulation, particularly under unconstrained lighting and limited views. Progress has also been limited by the lack of datasets that jointly provide accurate 3D geometry, high-resolution multi-view imagery, and a diverse population of subjects. To address this, we present PALM, a large-scale dataset comprising \numScans high-quality hand scans from 263 subjects and \numImages multi-view images, capturing rich variation in skin tone, age, and geometry. To show its utility, we present a baseline \methodname, a multi-subject prior over hand geometry and material properties learned via physically based inverse rendering, enabling realistic, relightable single-image hand avatar personalization. \datasetname's scale and diversity make it a valuable real-world resource for hand modeling and related research. See the project page at \href{https://github.com/facebookresearch/PALM}{this link}.

\blfootnote{$\dagger$ Work done during Meta internship}
\end{abstract}

\section{Introduction}
\label{sec:intro}

Human hands are central to how we interact with the physical and social world: we manipulate objects~\cite{hasson2019obman, wang2025magichoi,zhang2024artigrasp,fan2024benchmarks,fan2024hold}, express intent through gestures~\cite{yu2025dynhamr,fan2021learning,tse2023spectral}, and communicate affective cues via touch. Realistic and drivable hand avatars have the potential to transform virtual interaction, gaming, and telepresence. However, building such avatars from images remains a fundamentally challenging problem due to the complexity of hand geometry, appearance, and articulation, particularly under unconstrained lighting and from limited visual observations.

A critical missing component in the hand community is a large-scale, high-quality dataset that enables learning generalizable and physically grounded models of human hands. Existing datasets suffer from significant limitations: they often include only a small number of subjects~\cite{interhand,martinez2024codec}, lack accurate 3D hand geometry from real-world scans~\cite{interhand,mano}, or are derived from hand-crafted synthetic data~\cite{gao2022dart}, limiting their utility for learning models that generalize across identity and illumination.

To fill this gap, we introduce \datasetname, a large-scale dataset of human hands containing  publicly available-ready accurate hand scans, diverse in quantity and subject diversity. \datasetname includes \numScans high-quality 3D hand scans and \numImages high-resolution multi-view \rgb images from \numSubs subjects, each performing a diverse set of predefined hand poses designed to span a wide range of natural hand articulations. The subjects cover a broad range of skin tones and age groups. All data is captured using a commercial 3dMD scanner~\cite{3dmdhand}, providing precise, sub-millimeter geometry. Each scan is paired with synchronized multi-view images and a MANO registration (pose and shape), obtained via multi-view–consistent alignment to the 3D scans.  Importantly, the capture environment, including lighting and scanner configuration, remained fixed throughout the entire collection process, enabling consistent illumination conditions across subjects. 
While prior datasets have included either limited subjects or unreleased scan data, \datasetname will be made publicly available for research use upon publication, making it the most comprehensive and accessible dataset for studying generalizable hand models.

To highlight a practical use case for our data, we present a baseline model \methodname, an implicit neural prior over human hands that jointly models appearances, geometry and material properties using our dataset. \methodname is trained via physically based inverse rendering, decomposing each subject hand into geometry, albedo, specularity, roughness, and environment lights. The model conditions on pose and subject-specific latent codes, enabling it to capture pose-dependent effects. A key insight in \methodname is a shared environment lighting constraint across the subjects, which disentangles illumination from intrinsic hand appearance, allowing the model to generalize to novel lighting.

We apply our prior to the highly under-constrained task of single-image hand avatar personalization under unknown lighting. This monocular setting presents several challenges: depth ambiguity, occlusions (\eg, self-occlusion of the palm or dorsum), and ambiguous lighting. Our central insight is that, despite the fine-scale complexity of hands (\eg, wrinkles, creases, texture), many fundamental properties -- skin tone, material reflectance, and deformation behavior -- are shared across individuals and can be captured by a learned prior. By optimizing the subject-specific latent code and scene illumination, our model reconstructs realistic, relightable, and articulated hand avatars from a single input image, even in uncontrolled conditions.

To perform extensive evaluation of the single-image personalization setting, we evaluate our method in both synthetic and real-world datasets. 
Our results show that our method consistently outperforms other prior-based and non-prior-based methods for relightable hand personalization.

To summarize our contributions: 1) We introduce \datasetname, a large-scale dataset containing high-resolution \rgb multi-view images of diverse subjects with detailed hand scans and accurate MANO registrations; 2) 
We present a baseline, \methodname, a multi-subject implicit hand prior model that leverages \datasetname to learn physical hand properties such as geometry, albedo, specularity, and roughness;
3) We demonstrate the effectiveness of our prior model by using it to personalize and relight hand avatars from a single image under challenging and diverse environmental conditions.

\section{Related Work}
\label{sec:related_work}

\myparagraph{Hand datasets}
Recent years have seen an explosion of hand datasets, that can be categorized into hand-only~\cite{zimmermann2019freihand,moon2020deephandmesh,potamias2023handy,gao2022dart,martinez2024codec}, interacting hand~\cite{tzionas2016capturing,interhand,moon2023dataset}, and hand-object~\cite{hampali2020honnotate,fan2023arctic,dexycb,liu2022hoi4d,hasson2019obman,FirstPersonAction_CVPR2018,kwon2021h2o,contactpose_2020,sener2022assembly101,zhang2024graspxl} datasets. 
Early research on hand-only capture primarily focused on datasets collected using depth cameras~\cite{yuan2017bighand2}. Soon after, RGB-based datasets gained traction, with notable examples including STB~\cite{zhang20163d} and FreiHAND~\cite{zimmermann2019freihand}.
Following this, interest expanded toward hand-object interaction datasets. Hampali \etal~\cite{hampali2020honnotate} released a dataset of single hands manipulating rigid YCB objects, while Fan \etal~\cite{fan2023arctic} used a marker-based MoCap setup to capture full-body interactions with articulated objects. More recently, Banerjee \etal~\cite{banerjee2025hot3d} contributed a large-scale dataset featuring multi-object interactions.
Interacting-hand datasets have also grown in popularity. Tzionas \etal~\cite{tzionas2016capturing} pioneered capturing two-hand interactions using an RGB-D setup. Building on this, Moon \etal~\cite{interhand} leveraged a large-scale multi-view RGB system to recover 3D hand poses under strong self-contact, which significantly boosted interest in modeling two interacting hands from RGB data. Moon \etal~\cite{moon2023dataset} released a synthetic dataset using 3D annotations derived from~\cite{interhand}.
Handy~\cite{potamias2023handy} provides 3dMD data but it is not publicly available.
Despite the rapid emergence of various hand datasets, most approaches for learning hand avatars still rely primarily on InterHand2.6M~\cite{mundra2023livehand,chen2023handavatar} or on custom video recordings~\cite{karunratanakul2023harp}.
This is largely due to the absence of a high-quality dataset that simultaneously provides accurate 3D hand scans, high-resolution multi-view RGB imagery, and a diverse set of subjects.

\myparagraph{Hand representations}
Learning articulated hand representations is a long-standing research problem.
Some methods solely focus on modelling hand joints~\cite{fan2021learning,spurr2020eccv,zimmermann2017iccv} or geometries~\cite{huangPhrit, moon2020deephandmesh,mano,duran2024hmp,ziani2022tempclr}. 
For example, MANO~\cite{mano} pioneered parametric mesh-based hand geometry modeling, parameterizing shape with a PCA latent space. Moon \etal~\cite{moon2024authentic, moon2020deephandmesh} propose a non-linear approach for high-fidelity hand mesh modeling.  HALO~\cite{Korrawe2021halo} is an implicit articulated hand geometry representation using occupancy network via a differentiable canonicalization layer.
There are also methods that model both geometry and appearances of hands~\cite{potamias2023handy, qian2020html, chen2024urhand, li2022nimble, zheng2024ohta, corona2022lisa,mundra2023livehand,chen2023handavatar,fan2024hold}. 
HTML~\cite{qian2020html} extends MANO  with a PCA-based texture model. 
Handy~\cite{potamias2023handy} is a parametric hand model of shape and texture learned from proprietary hand scans. 
Both Handy and HTML preprocess texture maps to minimize baked-in shadows and specularities. 
LISA and OHTA~\cite{corona2022lisa} model shape and appearance fields, with lighting effects baked into the network and controlled by latent codes. 
NIMBLE~\cite{li2022nimble} models hands with bone, muscle, and skin deformation, using a light stage to capture pose-independent albedo and specular maps modeled with PCA bases. 
HARP~\cite{karunratanakul2023harp} optimizes the normal and albedo maps for the MANO hand mesh with a point light source to model shadow effects, demonstrating  slight generalizability to novel illuminations. 
URHand\cite{chen2024urhand} models pose-dependent hand material properties and is trained on large-scale light stage data. 
Capturing such pose-dependent material characteristics and lighting effects requires accurate environment map information, typically enabled by a sophisticated light stage setup with hundreds of synchronized cameras, as used in Nimble and URHand.
In contrast to these methods~\cite{corona2022lisa,zheng2024ohta,chen2024urhand}, our baseline method can jointly learn appearances, geometries and relight the hand avatar and does not rely on expensive light stage setup.

\section{\datasetname Dataset}
\label{sec:dataset}

\label{sec:dataset}
\myparagraph{Overview}
To study hand priors, we introduce \datasetname (see \refFig{fig:teaser}), a high-quality dataset with accurate 3D hand annotations, high-resolution multi-view RGB images, and 3dMD hand scans. It contains \numImages \rgb images and  \numScans hand scans from 263 subjects (131 male, 132 female) with diverse skin tones and hand shapes. Data was captured using a 3dMD hand scanner~\cite{3dmdhand} with 7 RGB and 14 monochrome machine vision cameras calibrated; the RGB images have a resolution of $2448 \times 2048$; and the 3D hand scans were reconstructed using 3dMD's software-driven triangulation technique based on Active Stereo Photogrammetry. Participants performed approximately 50 predefined right-hand gestures. See \suppl for more examples in \datasetname.


\subsection{Data Characteristics}

\begin{figure}[t]
        \centerline{\includegraphics[width=0.8\linewidth]{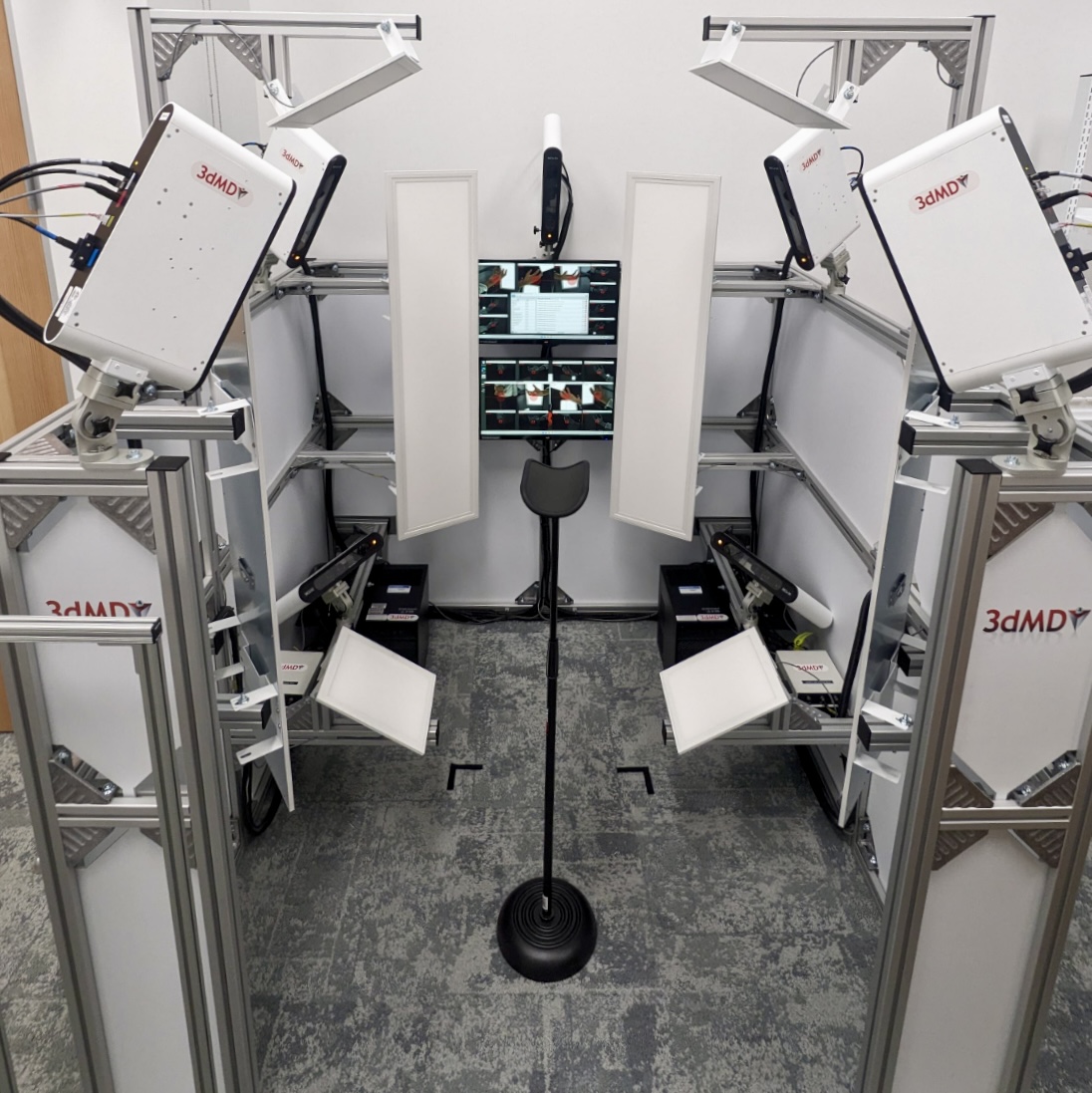}}
    \caption{
        \subtitle{Capture setup} Our 3dMD  setup with \numCams \rgb cameras.
    }
    \label{fig:dataset}
\end{figure}

\begin{table}[t]
\resizebox{1\columnwidth}{!}{
\begin{tabular}{ccccccc}
\toprule
dataset        & \# real sub. & \# scans & image size  & real RGB & annotation       & prior learning \\
\hline
ICVL~\cite{tang2014latent}           & 10               & 0        & $320 \times 240$   & -        & track            & $\times$ \\
BigHand2.2M~\cite{yuan2017bighand2}    & 10               & 0        & $640 \times 480$   & -        & marker           & $\times$ \\
Tzionas et al.~\cite{tzionas2016capturing} & -                & 0        & $640 \times 480$   & $\checkmark$      & track              & $\times$ \\
Simon et al.~\cite{simon2017hand}   & -                & 0        & $1920 \times 1080$ & $\checkmark$      & semi-auto        & $\times$ \\
EgoDexter~\cite{mueller2017real}      & 4                & 0        & $640 \times 480$   & $\checkmark$      & manual           & $\times$ \\
STB~\cite{zhang20163d}            & 1                & 0        & $640 \times 480$   & $\checkmark$      & manual           & $\times$ \\
FreiHAND~\cite{Freihand2019}       & 32               & 0        & $224 \times 224$   & $\checkmark$      & semi-auto        & $\times$ \\
InterHand2.6M~\cite{interhand}  & 50               & 0        & $512 \times 334$   & $\checkmark$      & semi-auto        & $\checkmark$ \\
Re:InterHand~\cite{moon2023dataset}   & 10               & 0        & $4096 \times 2668$ & $\times$       & -                & $\times$ \\
DART~\cite{gao2022dart}           & -                & 0        & $512 \times 512$   & $\times$       & -                & $\times$ \\
MANO~\cite{mano}           & 31               & 1K       & N/A         & $\checkmark$      & manual           & $\checkmark$ \\
\hline
PALM (Ours)    & 263              & 13K      & $2448 \times 2048$ & $\checkmark$      & semi-auto + scan & $\checkmark$ \\
\bottomrule
\end{tabular}
}
\caption{\subtitle{Publicly available datasets} Existing public datasets lack subject diversity, accurate hand scans, and high-quality multi-view \rgb images that are important for training a strong hand appearance and geometry prior model. Our dataset contains a large number of subjects with high-resolution images and scans suitable for learning a universal hand prior.}
\label{tab:datasets_comparison}
\end{table}

\myparagraph{Dataset comparison}
\refTab{tab:datasets_comparison} compares publicly available hand-only datasets. Early datasets~\cite{yuan2017bighand2, tzionas2016capturing, mueller2017real} focus on depth cameras and often have limited subjects and scale~\cite{tzionas2016capturing, mueller2017real}. Recent larger datasets focus on \rgb images: InterHand2.6M~\cite{interhand} has 50 subjects with multi-view calibrated \rgb, Re:InterHand~\cite{moon2023dataset} offers synthetically relit interacting-hand images of 10 subjects, and MANO~\cite{mano} provides hand scans of 31 subjects. Handy~\cite{potamias2023handy} includes 3dMD scans, but these are unavailable. While MANO, InterHand2.6M, and our dataset are all suitable for learning priors for hands, most other datasets are suboptimal due to missing modalities, limited subject diversity, synthetic data, or low image quality, leading most methods~\cite{moon2024authentic,mundra2023livehand,chen2023handavatar,karunratanakul2023harp} to rely on InterHand2.6M despite lacking scans. Our dataset is substantially larger in scale, with \numSubs subjects, real high-resolution calibrated \rgb images, and \numScans  high-quality scans, making it ideally suited for learning robust hand priors; it will be released publicly to advance future research.

\begin{figure}[t]
    \centerline{\includegraphics[width=1\linewidth]{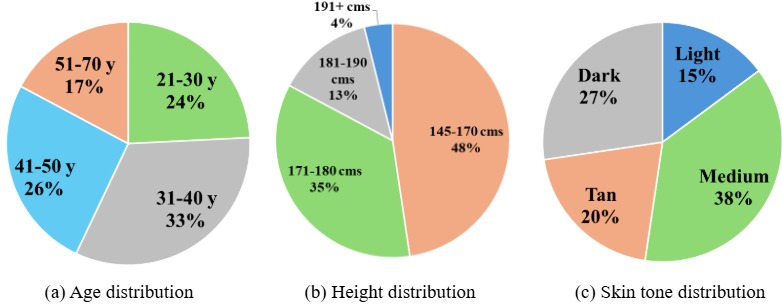}}
    \caption{
        \subtitle{\datasetname demographics} (a) Age; (b)  Height; (c) Skin tone distributions. Our dataset provides a wide distribution of skin tones and age groups representing a large variety of hand textures.
    }
    \label{fig:demographics}
\end{figure}

\myparagraph{Demographics}
\refFig{fig:demographics} provides a detailed breakdown of the demographic distribution in \datasetname. The dataset includes participants aged 21--70 years, with the majority in the 31--40 (33\%), 41--50 (26\%), and 21--30 (24\%) age brackets.
In terms of height, subjects range from 145 to 200\,cm, with 48\% in the 145--170\,cm range and 35\% in the 171--180\,cm range. Only a small portion (4\%) are taller than 190\,cm.
Skin tone distribution is also diverse, comprising 38\% medium, 27\% dark, 20\% tan, and 15\% light tones. This diversity supports robust analysis across demographic variations.

\subsection{Data Acquisition}

\myparagraph{Capture setup}
We use a 3dMD~\cite{3dmdhand} hand scanner to capture high-resolution multi-view RGB images and 3D scans of the subjects. In particular, the 7-viewpoint setup consists of an array of 21 synchronized and calibrated machine vision cameras, including both RGB and monochrome sensors, arranged to provide full 360-degree coverage of the hand. The system captures images at a high resolution of $2448 \times 2048$. Random light projectors are integrated to enhance surface detail and geometry accuracy. The 3D hand scans were reconstructed using 3dMD's software-driven triangulation technique based on Active Stereo Photogrammetry. All cameras are calibrated, ensuring consistent alignment across views. This setup enables the acquisition of dense, accurate 3D hand meshes in various poses, suitable for studying both static poses and dynamic hand articulations.

\myparagraph{Capture protocol}  
Each participant is asked to stand still with their right hand placed inside the 3dMD capture volume.  
Participants are instructed to perform approximately 50 predefined right-hand gestures, covering a wide range of articulations including open-hand poses, pinches, fist closures, and fine-grained finger movements.  
To ensure consistency across subjects, a standardized gesture list is followed, and participants are guided through the sequence during the capture.  
We capture each gesture as an independent hand scan.
This protocol ensures diverse and repeatable motion data across the entire subject pool.
All subjects are captured in the same lighting and camera setup.

\myparagraph{3D keypoint annotation}
A semi-automatic pipeline is used to generate accurate \threeD hand pose labels. The \twoD keypoints are first manually annotated on a subset of images to train a \twoD keypoint detector tailored to our capture setup. The detector, implemented as a U-Net~\cite{ronneberger2015u} pre-trained on InterHand2.6M and fine-tuned on 10K manually annotated \datasetname images, is then used to estimate \twoD keypoints for all camera views, which are subsequently triangulated to obtain \threeD poses using multi-view geometry.
Specifically, we follow the approach used in InterHand2.6M and apply RANSAC-based triangulation to robustly solve for the \threeD keypoint locations. This semi-automatic approach significantly reduces the manual labeling burden while ensuring reliable \threeD pose annotations.

\myparagraph{MANO registration}
The 3D MANO poses for each pose and subject are obtained by registering the MANO hand model to the hand scans. To register MANO to each hand gesture, we optimize the MANO hand model using a combination of 2D/3D keypoints, segmentation mask, and 3D hand scan supervision. Specifically, we minimize the closest-point distance from each MANO vertex to the corresponding hand scan surface.
Ground-truth masks are derived from the hand scans, and we employ Soft Rasterizer~\cite{liu2019soft} for differentiable rendering to align the MANO silhouette with these masks. 

The registration is independently performed for each subject in two stages. The first stage optimizes the per-subject hand shape parameters and per-frame pose parameters on a set of simple hand poses (\eg, flat hand), and the second stage only optimizes the per-frame pose parameters while freezing the shape parameters. Our final \threeD keypoints have a recall rate of 95\% at 10mm threshold. Our MANO registrations show a mean fitting error of 5.3 mm with respect to the \threeD keypoints (similar to that of InterHand2.6M~\cite{interhand}).

\section{Method}
\label{sec:method}

\methodname, illustrated in \refFig{fig:method}, utilizes \datasetname, our large-scale collection of human hand data containing detailed hand scans and high-resolution \rgb images, to train a \textit{personalizable} and \textit{relightable} hand prior.
In this section, we first introduce preliminary concepts on NeRF~\cite{mildenhall2021nerf}, the Neural Radiance Field technique at the core of our approach, and \mano\ccite{mano}, the parametric hand model that we leverage within our pipeline to map 3D points across different subjects and hand poses into a shared \textit{canonical} representation. Next, we present \methodname, our novel framework for learning a multi-subject hand prior through physically based inverse rendering (PBR), and describe how to train our representation over multiple subjects. Finally, we detail how \methodname can be used to recover a \textit{personalized} and \textit{relightable} hand avatar from a single image. {This is a highly under-constrained problem, and we show that a prior model such as \methodname helps in recovering realistic, personalized hand avatars even in extreme illumination settings.

\subsection{Preliminaries}

\myparagraph{NeRF} Given a ray $\mathbf{r} = ( \mathbf{o}, \mathbf{d} )$ defined by its camera
center $\mathbf{o}$ and viewing direction $\mathbf{d}$, NeRF \cite{mildenhall2021nerf} computes the output
radiance (\ie, pixel color) of the ray via:
\begin{align}
\label{eq:nerf_exact}
  C_{rf}(\mathbf{r}) =& \int_{t_n}^{t_f} T(t_n, t) \sigma_t (\mathbf{r}(t)) L (\mathbf{r}(t), \mathbf{d}) dt \\
  \text{s.t} \quad & \mathbf{r} (t) = \mathbf{o} + t \mathbf{d} \nonumber \\
  & T(t_n, t) = \exp \left( - \int_{t_n}^t \sigma_t(\mathbf{r}(s)) ds \right ) \nonumber
\end{align}
where $t_n, t_f$ denote the near/far point for the ray integral; $\sigma_t(\mathbf x): \mathbb R ^{3} \rightarrow \mathbb R $ is a neural network that models surface density at 3D point $\mathbf x$ ; $L(\mathbf x, \mathbf{d}): \mathbb R ^{3} \times \mathbb R ^{3} \rightarrow \mathbb R $ is a neural network that parametrizes radiance color at 3D point $\mathbf x$ when observed from direction $\mathbf{d}$. In practice, the integrals above are approximated via quadrature, yielding:
\begin{align}
\label{eq:nerf_approx}
  C_{rf}(\mathbf{r}) \approx & \sum_{i=1}^{N} w^{(i)} L (\mathbf{r}(t^{(i)}), \mathbf{d}) \\
  \text{s.t} \quad & \mathbf{r} (t) = \mathbf{o} + t \mathbf{d} \nonumber \\
  & w^{(i)} = T^{(i)} \left(1 - \exp(-\sigma_t(\mathbf{r}(t^{(i)})) \delta^{(i)} \right) \nonumber \\
  & T^{(i)} = \exp \left( -\sum_{j < i} \sigma_t(\mathbf{r}(t^{(j)})) \delta^{(j)} \right) \nonumber \\
  & \delta^{(i)} = t^{(i+1)} - t^{(i)} \nonumber,
\end{align}
where $\{ t^{(1)}, \cdots, t^{(N)} \}$ are a set of sampled points on the ray that are obtained through importance sampling and $\delta^{(i)}$ is the length of the $i^{\text{th}}$ sampling interval.

\paragraph{\mano and canonical representation.}
The \mano\ccite{mano} hand model is parametrized by $\V{\Theta} = \{ \V{\theta}, \V{\beta}, \V{p}\}$, where $\V{\theta} \in \R^{45}$ denotes hand skeletal pose (joint angles), $\V{\beta}\in \R^{10}$ hand shape (parameterized by PCA coefficients) and $\V{p} \in \R^{6}$ global transformation.
The \mano model then maps $\V{\Theta}$ to a posed \threeD mesh $\mathcal{M}(\V{\Theta})\in \R^{778\times 3}$.
\methodname leverages \mano to perform inverse LBS using SNARF~\cite{chen2021snarf} that map points in \threeD to a common canonical representation, i.e., given a 3D point in \textit{deformed} coordinates $\mathbf x_d$, hand parameters $\V{\Theta}$, we find the corresponding 3D point $\mathbf x_c$ in \textit{canonical} space as,
\begin{equation}
    \mathbf x_d = \argmin_{\mathbf x} \Big\lvert \Big\lvert \sum_{i=1}^{n_b} w_i(\mathbf x)\cdot B_i \cdot \mathbf x - \mathbf x_d\Big\lvert \Big\lvert_2^2,
\end{equation}
where, $w_i(\mathbf x)$ is the skinning weight associated with point $\mathbf x$ for bone $i$, and $B_i$ represents the $i^{\text{th}}$ bone transformation.

\begin{figure*}[t]
    \centerline{\includegraphics[width=1\linewidth]{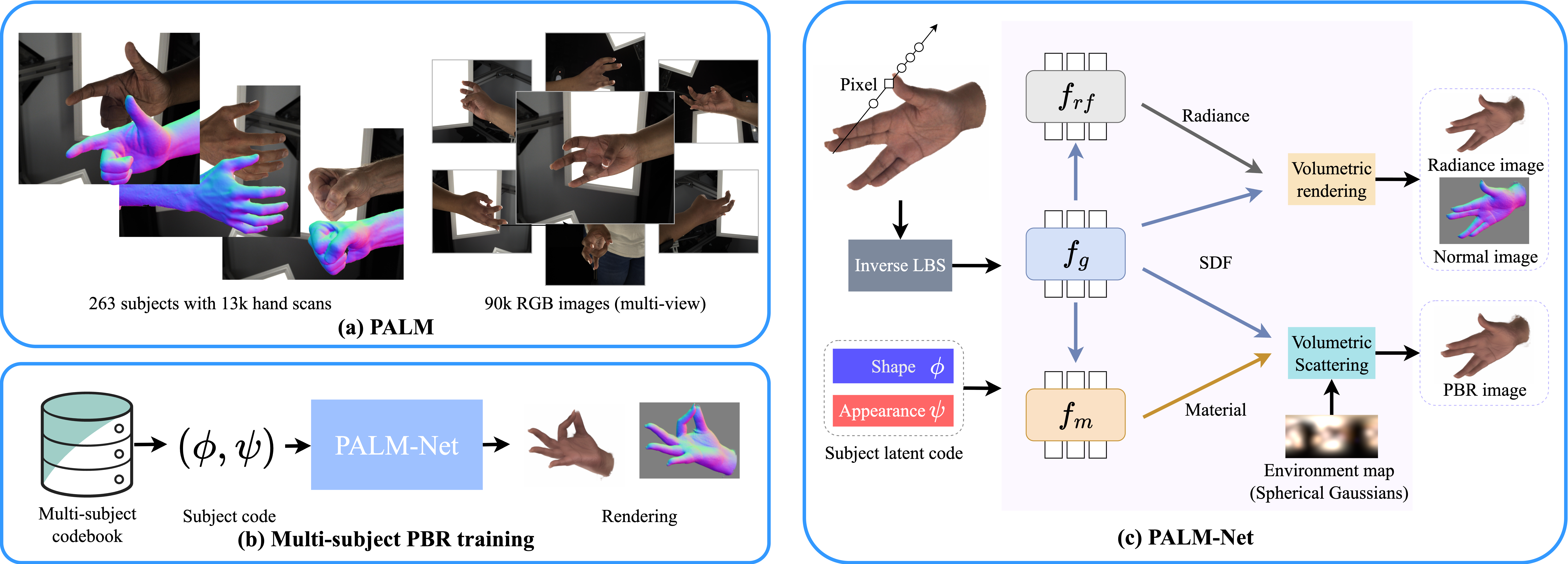}}
    \caption{
        \subtitle{\methodname overview} Given (a) \datasetname, our multi-subject \rgb dataset with \numSubs subjects, \methodname explains each subject by optimizing subject-specific shape and appearance codes (b). (c) \methodname is an implicit physically-based network that is conditioned on the subject codes and renders to radiance, normal, and physically-based \rgb images. 
    }
    \label{fig:method}
\end{figure*}

\subsection{\methodname}

\myparagraph{Physically based representations}
Inspired by~\cite{wang2024intrinsicavatar,chen2022gdna}, \methodname (\refFig{fig:method}c) decomposes the hand representation into a shape network $f_{g}(\cdot)$, a radiance field network $f_{rf}(\cdot)$, and a material network $f_{m}(\cdot)$. We model the canonical hand geometry with an implicit function:
\begin{align}
    f_g: \mathbf x_c, \theta, \beta,  \phi &\mapsto \sigma_t(\mathbf{x}_c), z(\mathbf{x}_c),
\end{align}
where $\mathbf x_c \in \R^3$ denotes a 3D point in the canonical space, $\theta, \beta$ are \mano parameter, and $\phi \in \R ^{d_s}$ captures subject-specific geometry latent code of dimension $d_s$.
$f_g(\cdot)$ outputs the opacity value, $\sigma_t(\mathbf{x}_c) \in \R$, as well as geometry features $z(\mathbf{x}_c) \in \R ^{d_s}$ for the point $\mathbf x_c$.
Following~\cite{yariv2021volume}, the opacity is obtained by converting the Signed Distance Function~(SDF) values via the cumulative distribution function of the scaled Laplace distribution, $\Gamma_{\alpha_1, \alpha_2}(s)$, where $\alpha_1, \alpha_2 >0$ are optimizable parameters. For~details, we refer the reader to~\cite{yariv2021volume}. 

The outgoing radiance, $L(\mathbf x_c, \mathbf{d})$ at canonical point $x_c$ viewed by the direction $\mathbf{d} \in \R^3$ is obtained as,
\begin{align}
    f_{rf}: \mathbf x_c, z, \text{ref}(\mathbf d, \mathbf n), \mathbf n, \theta, \psi &\mapsto L(\mathbf x_c, \mathbf{d}),
\end{align}
where, $\mathbf{n}$ is the surface normal obtained analytically from the SDF field, $\text{ref}(\mathbf d, \mathbf n)$ reflects the view direction $\mathbf{d}$ around the normal $\mathbf n$, and $\psi \in \R^{d_s}$ is the appearance latent code.

Lastly, the spatially varying material field, $f_m$  is used to model the physically based rendering parameters, the albedo $\alpha \in \R^3$, roughness $r \in \R$, and metallicity $m \in \R$ as, 
\begin{align}
    f_m: \mathbf x_c, z, \theta, \psi &\mapsto \alpha(\mathbf x_c), r(\mathbf x_c), m(\mathbf x_c)
\end{align}

The canonical point $\mathbf x_c$ is encoded with hash grid encoding for all three networks, $f_{\{g,~rf,~m\}}$ to model high frequency details efficiently.

\myparagraph{Physically based rendering}
For physically based rendering, we follow closely \cite{wang2024intrinsicavatar} and compute the radiance scattered by the volume along a certain camera ray ($\mathbf{o,d}$) using the quadrature approximation as,
\begin{align}
\label{eq:pbr_approx}
  C_{pbr}(\mathbf{r}) &\approx \sum_{i=1}^M w^{(i)} BRDF\Big(\mathbf{d}, \bar{\mathbf{d}}^{(i)}, \alpha\big(\mathbf{r}(\bar{t}^{(i)})\big), r\big(\mathbf{r}(\bar{t}^{(i)})\big),\nonumber\\ &~~~~~~~~~~~~~~m\big(\mathbf{r}(\bar{t}^{(i)})\big), \mathbf{n}\Big) \cdot L_i (\mathbf{r}(\bar{t}^{(i)}), \bar{\mathbf{d}}^{(i)}) \cdot \frac{1}{pdf(\bar{\mathbf{d}^{(i}})} \\
  \text{s.t} &\quad  \mathbf{r} (t) = \mathbf{o} + t \mathbf{d} \nonumber \\
  & w^{(i)} = T^{(i)} \left(1 - \exp(-\sigma_t(\mathbf{r}(\bar{t}^{(i)}) \delta^{(i)} ) \right) \nonumber \\
  & T^{(i)} = \exp \left( -\sum_{j < i} \sigma_t(\mathbf{r}(\bar{t}^{(j)})) \delta^{(j)} \right) \nonumber \\
  & \delta^{(i)} = \bar{t}^{(i+1)} - \bar{t}^{(i)} \nonumber
\end{align}
where $(\bar{t}^{(1)}, \bar{t}^{(2)},..., \bar{t}^{(M)})$ are the importance sampling offsets from the PDF estimated by radiance field samples in \refEq{eq:nerf_approx}; $M$ denotes the number of samples used to approximate the integrals along the ray; $\bar{\mathbf{d}}^{(i)}$ is the incoming light direction at sampling offset $\bar{t}^{(i)}$ sampled from the distribution, $pdf(\cdot)$ (uniform distribution over the unit sphere); $BRDF(\cdot)$ denotes the simplified version of Disney BRDF~\cite{Burley2012SIGGRAPH}. 
The term $L_i(\mathbf{x}, \bar{\mathbf{d}})$ is the incoming radiance towards point $\mathbf{x}$ along direction $\bar{\mathbf{d}}$ and can be computed as the weighted sum of output radiance $C_{rf}(\mathbf x , \bar{\mathbf{d}})$ (\refEq{eq:nerf_approx}) and radiance emitted from an environment map $\text{Env}(\bar{\mathbf{d}})$:
\begin{align}
L_i(\mathbf{x}, \bar{\mathbf{d}}) = &C_{rf}(\mathbf x , \bar{\mathbf{d}}) \\ + &\exp \left( - \int_{t_n'}^{t_f'} \sigma_t (\mathbf x + s \bar{\mathbf d}) ds \right) \text{Env} (\bar{\mathbf d}),
\end{align}
where ${t_n'}, {t_f'}$ are near and far points of integration for secondary rays. The environment map is approximated by a set of Spherical Gaussians denoted by $\mathcal{SG}_1, \mathcal{SG}_2,...,\mathcal{SG}_G$. We refer the reader to \cite{wang2024intrinsicavatar} for derivation of the above equations.

\myparagraph{Training losses}
Since reconstructing geometries and material properties from \rgb images is a highly under-constrained problem, we devise a loss $\mathcal{L}$ that consists of several terms. 
 In particular, we first encourage \rgb values to be consistent with an input image via 
\begin{align}
\mathcal{L}_\text{rf} =  \sum_{\textbf{r}} \norm{C_{rf}(\V{r}) - \hat{C}(\V{r})},
\end{align}
where $\V{r}$ is a ray casted from a sampled pixel on an image, and $C_{rf}(\V{r})$ and $\hat{C}(\V{r})$ are the rendered radiance value and \groundtruth color.
The PBR rendered pixels $C_{pbr}$ are directly supervised with RGB values in a loss  $\mathcal{L}_\text{pbr}$ similar to $\mathcal{L}_\text{rf}$. 
Since our  scans provide detailed  geometries, we supervise the model with the rendered scan normals:
\begin{align}
    \mathcal{L}_\text{normal} = \sum_{\textbf{r}} \norm{\mathcal{N}(\V{r}) - \hat{\mathcal{N}}(\V{r})}.
\end{align}
Note that the geometry information is shared by PBR and radiance field.
We encourage valid SDFs with the eikonal loss $\mathcal{L}_\text{eikonal}$~\cite{gropp2020igr}, which enforces the gradient at each point to have a unit norm.
To encourage smooth hand surfaces, we apply a Laplacian loss $\mathcal{L}_\text{LAP}$ on sampled points around the hand (see \suppl).
To encourage latent codes to be close to zeros, we penalize a MSE loss $\mathcal{L}_\text{latent} $ on both appearance and shape code. 
To avoid foreground model explaining background pixels, we also supervise the networks with a segmentation loss 
\begin{align}
    \mathcal{L}_\text{segm} = \sum_{\textbf{r}} \text{BCE}(\mathcal{S}(\V{r}), \hat{\mathcal{S}}(\V{r}))
\end{align}
where $\mathcal{S}(\V{r}) \in \R$ represents the probability of a pixel being the foreground and $\text{BCE}(\cdot, \cdot)$ is the binary cross entropy loss to the \groundtruth hand segmentation mask $\hat{\mathcal{S}}(\V{r})$ rendered from the hand scans. 
Finally, to capture high-frequency details, we render image patches and compare them with the \groundtruth using the perceptual similarity loss $\mathcal{L}_\text{LPIPS}$~\cite{zhang2018perceptual}.
The total loss $\mathcal{L}$ is defined as  
\begin{align}
\mathcal{L} =
     \mathcal{L}_\text{rf} &+ \lambda_\text{pbr}\mathcal{L}_\text{pbr} + 
     \lambda_\text{segm}\mathcal{L}_\text{segm} + \lambda_\text{normal}\mathcal{L}_\text{normal} \notag \\ 
     &+ 
    \lambda_\text{eikonal}\mathcal{L}_\text{eikonal} + \lambda_\text{LPIPS}\mathcal{L}_\text{LPIPS} \\
    &+ \lambda_\text{LAP}\mathcal{L}_\text{LAP} + \lambda_\text{latent}\mathcal{L}_\text{latent} 
\end{align}
where $\lambda_{*}$ are the weights for the loss terms (see \suppl). 
We gradually decrease $\lambda_\text{segm}$ over time.

\myparagraph{Multi-subject PBR prior}
When training \methodname across multiple subjects, 
we model the detailed hand shape and appearance of each subject by conditioning \methodname on a shape code $\phi$ and an appearance code $\psi$. 
That is, we model subject identities by disentangling shapes from appearances. Empirically, we found that when training on multiple subjects, having separate latent codes for geometries and appearances yields better reconstructions than having a shared latent code for both.

Given a set of images of hands $\{\mathcal{I}\}$ from $N$ subjects, and randomly initialized latent codes $\{\phi_{1..N}, \psi_{1..N}\}$, \methodname explains the images of multiple subjects by optimizing on the network weights $\Phi$, the subject codebook  $\{\phi_{1..N}, \psi_{1..N}\}$, and Spherical Gaussian parameters for the environment lights.
Note that we optimize for a single environment across all subjects as the subjects are  captured in the same setup. 
In particular, our objective function is,
\begin{align}
    \min_{\Phi, \{\phi_i\}_{i=1}^N, \{\psi_i\}_{i=1}^N, \{\mathcal{SG}_i\}_{i=1}^G}  \mathcal{L}
\end{align}

\myparagraph{Personalization}
A strong prior model on the hand appearance and geometry allows us to personalize our model to images of hands captured in extreme environment settings. This is mainly because the prior model constrains the albedo and material properties of the hand during personalization and all the environment effects could be explained separately. This is achieved by solving an optimization problem where the shape code $\phi$ and the appearance code $\psi$ for a given input image are optimized along with the environment map while keeping the network weights $\Phi$ of the PBR prior model frozen. In particular, at each iteration, we sample a random batch of rays from the input image and optimize for the following objective:
\begin{align}
    \min_{\phi, \psi, \{\mathcal{SG}\}_{i=1}^G}      \mathcal{L}_\text{rf} &+ \lambda_\text{pbr}\mathcal{L}_\text{pbr} + 
     \lambda_\text{segm}\mathcal{L}_\text{segm} \nonumber\\ 
     &+ \lambda_\text{LPIPS}\mathcal{L}_\text{LPIPS} 
\end{align}

\section{Experiments}
\label{sec:exp}
\begin{figure}[t]
    \centerline{\includegraphics[width=\linewidth]{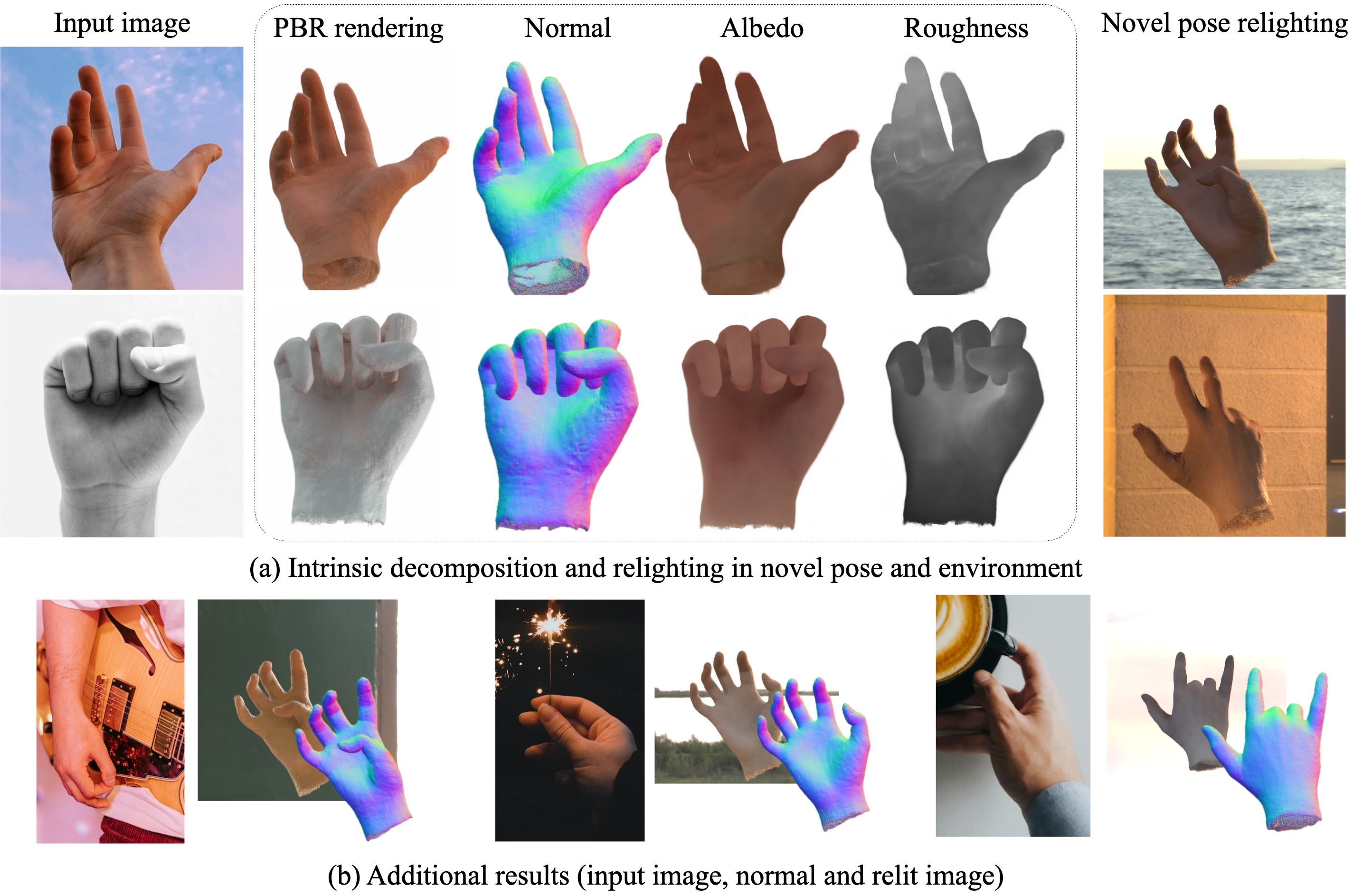}}
    \captionof{figure}{\subtitle{In-the-wild image personalization} 
    (a) The first column shows the images used for personalization, followed by the renderings of the geometry and materials of the hand avatar obtained using our prior model. The PBR rendering refers to the physically-based rendering with estimated environment map. The last column shows the relighting results of personalized hand avatar in a novel pose. Our method retrieves realistic hand avatars even when the input personalization image has complex lighting effects. (b) Additional relighting results with in-the-wild images.
    }
    \label{fig:itw_qualitative}
\end{figure}%

\noindent We evaluate our baseline on the task of hand avatar personalization and relighting from a single RGB image using three different datasets.
Metric and baseline details are in \suppl.

\subsection{Datasets}
\myparagraph{InterHand2.6M}
To evaluate the performance on real images, we use images from InterHand2.6M~\cite{interhand} for personalization. 
The dataset consists of accurate \threeD hand poses of subjects in predefined poses as well as high-resolution \rgb images. 
We evaluate on right-hand only sequences using the test split of the dataset. 
To reliably measure performance, we randomly select two views for each sequence, resulting in 12 sequences. 
We uniformly sample 20 images for each sequence for the evaluation and use the first frame of each sequence for training a personalized model.

\myparagraph{HARP relit} 
Since there is no real dataset to evaluate hand avatar under novel environment and poses, following~\cite{karunratanakul2023harp}, we render a synthetic dataset using Blender.
In particular, to create synthetic hand template, for each sequence, we sample new MANO shape parameters and apply a new skin tone using UV textures from DART~\cite{gao2022dart}. 
To animate the hand, we use the hand pose parameters from HARP data release~\cite{karunratanakul2023harp}.
We use the first frame of each sequence for training a personalized model and use the remaining frames for evaluation. We render the training and evaluation images using different environment maps to evaluate relighting.

\myparagraph{In-the-wild images} 
To show the generalization of our method under real world scenarios, we select in-the-wild images from the internet with diverse lighting conditions and poses. After personalization on these images, we render them with novel environment maps using novel poses from~\cite{karunratanakul2023harp} and show qualitative results.

\begin{table}[t] 
\centering
\setlength{\tabcolsep}{7mm}{
\resizebox{1.0\columnwidth}{!}{
\centering
\begin{tabular}{lrrr}
\toprule 
Method & PSNR$\uparrow$ & SSIM$\uparrow$ & LPIPS$\downarrow$ \\ 
\midrule
Handy~\cite{potamias2023handy} & 7.50 & 0.69 & 0.24 \\
HARP~\cite{karunratanakul2023harp} & 9.89 & 0.78 & 0.16 \\
UHM~\cite{moon2024authentic} & 10.08 & 0.76 & 0.19 \\
Ours & \textbf{12.01} & \textbf{0.84} & \textbf{0.15} \\ 
\bottomrule
\end{tabular}
}} 
\caption{\subtitle{\interhand dataset evaluation} Comparison of methods on single-image personalization task using PSNR, SSIM, and LPIPS metrics. Our method outperforms previous methods in the novel pose setting where the training and evaluation environment maps are the same.}
\label{tab:ih_exp_summary}
\end{table} 

\begin{table}[t] 
\centering
\setlength{\tabcolsep}{7mm}{
\resizebox{1.0\columnwidth}{!}{
\centering
\begin{tabular}{lrrr}
\toprule 
Method & PSNR$\uparrow$ & SSIM$\uparrow$ & LPIPS$\downarrow$ \\ 
\midrule
Handy~\cite{potamias2023handy} & 12.48 & 0.76 & 0.32 \\
HARP~\cite{karunratanakul2023harp} & 11.93 & 0.69 & 0.37 \\
UHM~\cite{moon2024authentic} & 12.30 & 0.74 & \textbf{0.31} \\
Ours & \textbf{13.39} & \textbf{0.78} & 0.35 \\ 
\bottomrule
\end{tabular}
}} 
\caption{\subtitle{Synthetic dataset evaluation} Comparison of methods based on PSNR, SSIM, and LPIPS. Our method outperforms previous methods in the novel environment, novel pose setting showing that the appearance reconstructions of our model is more accurate.}
\label{tab:syn_exp_summary}
\end{table}

\begin{figure*}[t]
    \centerline{\includegraphics[width=0.8\linewidth]{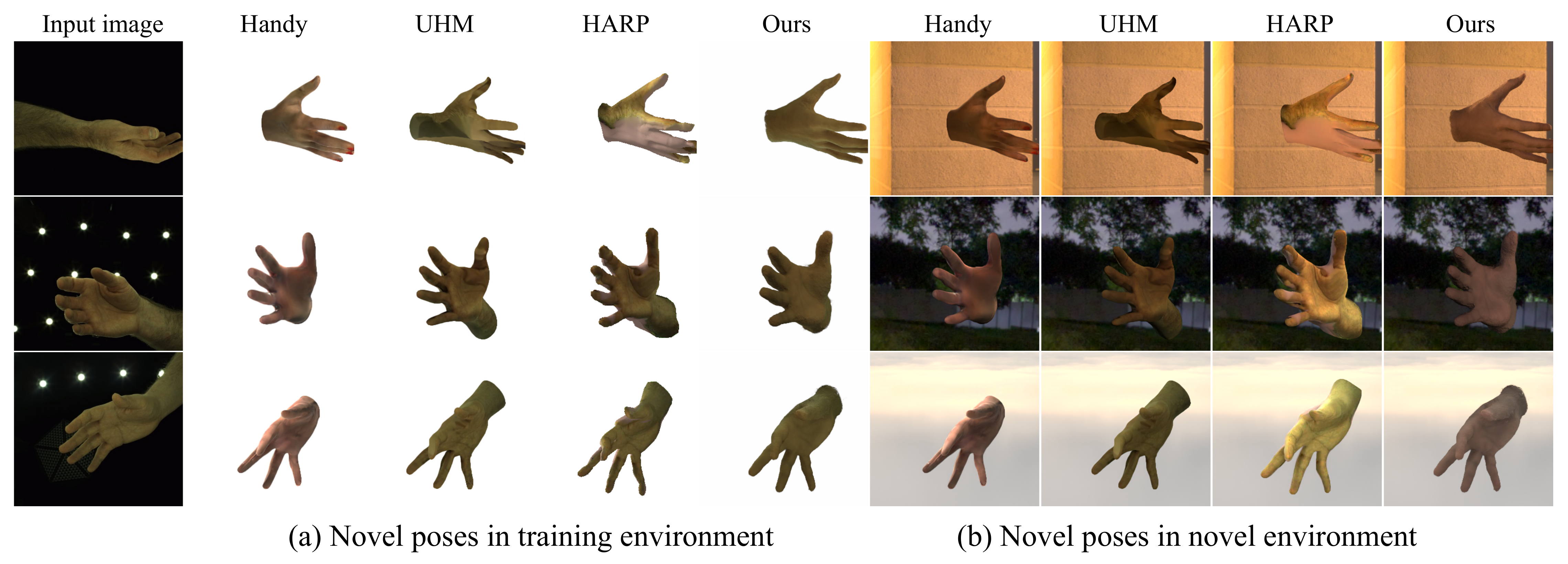}}
    \captionof{figure}{
        \subtitle{Personalization results on InterHand2.6M} The first column shows the image used for personalization. (a) Personalized hand avatars rendered in novel poses in the training environment. (b) Personalized hand avatars rendered in novel poses in the novel environment. The hand avatars from our method are more realistic than other baselines.
    }
    \label{fig:ih_qualitative}
\end{figure*}

\subsection{Comparison and Analysis}

\myparagraph{Baseline comparison} Table~\ref{tab:ih_exp_summary} compares the performance of our method with the baselines on \interhand. Our method outperforms baseline methods on all metrics showing high quality rendering in novel poses and viewpoints. Figure \ref{fig:ih_qualitative} shows the qualitative images on \interhand dataset for both the training and novel environments. The images from \methodname are more realistic than that of the baselines.
\refTab{tab:syn_exp_summary} compares  ours with the baselines on the synthetic dataset, where the evaluations are performed in novel environment settings. Our \methodname prior model outperforms the baselines by showing more realistic relighting results in novel environments (more results in \suppl).
\refFig{fig:itw_qualitative} shows qualitative results of personalization on in-the-wild images with diverse lighting condition and poses. Despite these challenges, our method produces realistic avatars and plausible relit images. 
For example, even in extreme setting where the input image is grayscale, our method can still recover plausible albedo thanks to our prior on hand appearance and the optimization over the environment map.

\begin{table}[t]
\centering
\begin{tabular}{cccc}
\toprule
\multicolumn{1}{l}{} & PSNR  & SSIM & LPIPS \\ \hline
w/o normal          & 11.97 & 0.84 & 0.18 \\ \hline
w/ normal           & \textbf{12.01} & \textbf{0.84} & \textbf{0.15} \\
\bottomrule
\end{tabular}
\caption{\subtitle{Effects of 3dMD normals}}
\vspace{-2mm}
\label{tab:ablate_normal}
\end{table}

\begin{figure}[t]
    \centerline{\includegraphics[width=0.7\linewidth]{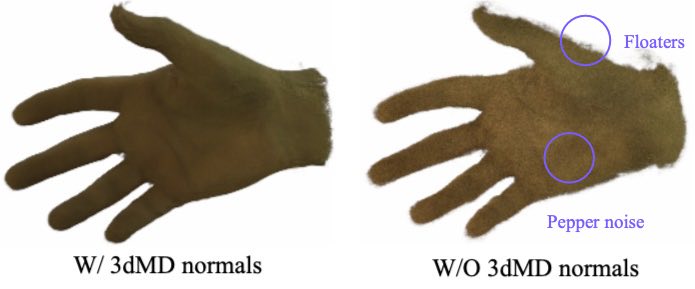}}
    \captionof{figure}{\textbf{Effects of 3dMD scans for hand personalization.} 
    }
    \vspace{-2mm}
    \label{fig:ablate_normal}
\end{figure}%

\myparagraph{Supervision with 3dMD normals}
Traditional multi-view techniques require a massive amount of \rgb views for high-fidelity 3D reconstruction~\cite{mildenhall2021nerf}. 
In our capture setup, we use a hybrid approach, combining 3dMD scans and sparse multi-view \rgb images for training our prior models. 
\refFig{fig:ablate_normal} shows a qualitative comparison with and without 3dMD normals. 
In particular, we train two multi-subject PBR prior models on \datasetname, one with 3dMD normal supervision and one without. 
Then we take these two prior models and personalize on an InterHand2.6M image. 
\refFig{fig:ablate_normal} shows that, 3dMD normals from hand scans are crucial in reducing pepper-like artifacts and it helps to reduce floaters. 
\refTab{tab:ablate_normal} quantitatively compares in this novel pose evaluation.

\begin{table}[t]
\centering
\begin{tabular}{cccc}
\toprule
\multicolumn{1}{l}{} & PSNR  & SSIM & LPIPS \\ \hline
w/o env          & 16.14 & 0.79 & 0.229 \\ \hline
w/ env           & \textbf{16.74} & \textbf{0.81} & \textbf{0.222} \\
\bottomrule
\end{tabular}
\caption{
\subtitle{Effects of modelling environment lightings}
}
\vspace{-2mm}
\label{tab:env_light}
\end{table}

\begin{figure}[t]
    \centerline{\includegraphics[width=0.7\linewidth]{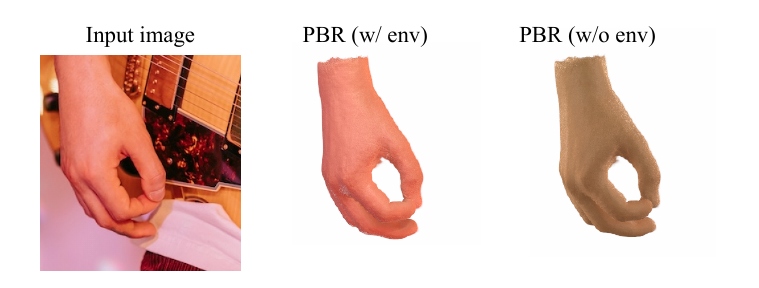}}
    \caption{
        \subtitle{The effect of modelling environment} 
    }
    \vspace{-4mm}
    \label{fig:env_map}
\end{figure}

\myparagraph{Environment map optimization}
During personalization, \methodname explains the image with  geometry, material properties and environment lightings. 
\refTab{tab:env_light} shows that modelling environment lightings enables the model to be more expressive in fitting the input image. An example of physically based rendered \rgb images are show in \refFig{fig:env_map}. When allowed optimizing the environment in \methodname, the fitting results are closer to that of the input image.

\section{Conclusion}
\label{sec:conclusion}

PALM is a large-scale dataset combining accurate 3D hand geometry, high-resolution multi-view imagery, and a diverse subject pool, addressing key limitations in existing datasets. Through \methodname, we demonstrate that physically based inverse rendering with a multi-subject prior enables realistic, relightable single-image personalization. The dataset's scale, diversity, and accompanying baseline make it a solid resource for future work.

{
    \small
    \bibliographystyle{ieeenat_fullname}
    \bibliography{main}
}

\end{document}